\documentclass[11pt]{article}

\usepackage[final]{acl}
\usepackage{times}
\usepackage{latexsym}
\usepackage[T1]{fontenc}
\usepackage[utf8]{inputenc}
\usepackage{microtype}
\usepackage{inconsolata}
\usepackage{graphicx}
\usepackage{xurl}
\usepackage{amsmath}
\usepackage{booktabs}

\title{GATech at AbjadGenEval Shared Task: Multilingual Embeddings for Arabic Machine-Generated Text Classification}

\author{Ahmed Khaled Khamis \\
  Georgia Institute of Technology \\
  \texttt{akhamis6@gatech.edu}}

\date{}

\begin{document}
\maketitle

\begin{abstract}
We present our approach to the AbjadGenEval shared task on detecting AI-generated Arabic text. We fine-tuned the multilingual E5-large encoder for binary classification, and we explored several pooling strategies to pool token representations, including weighted layer pooling, multi-head attention pooling, and gated fusion. Interestingly, none of these outperformed simple mean pooling, which achieved an F1 of 0.75 on the test set. We believe this is because complex pooling methods introduce additional parameters that need more data to train properly, whereas mean pooling offers a stable baseline that generalizes well even with limited examples. We also observe a clear pattern in the data: human-written texts tend to be significantly longer than machine-generated ones.
\end{abstract}

\section{Introduction}
When ChatGPT \cite{gpt4} and similar models started producing fluent Arabic text, it became clear that detection tools would need to catch up. Unlike English, where several detectors already exist, Arabic has received far less attention, mostly due to its morphological complexity and the diversity of written styles across regions. AbjadGenEval \cite{abudalfa-etal-2025-arageneval} \cite{ezzini2026abjadgeneval} \cite{daigt} addresses this gap with a shared task specifically for Arabic machine-generated text detection.

We approached this as a classification problem: take a pre-trained multilingual encoder \cite{transformers} (E5-large) \cite{e5}, add a classification head, and fine-tune on the provided data. The interesting part was figuring out how to pool the token representations. We implemented weighted layer pooling (learning which transformer layers matter most), attention-based pooling (learning which tokens to focus on), and gated fusion (learning how to combine multiple pooling outputs). After all that engineering, plain mean pooling gave us the best results.

This paper makes three contributions:
\begin{itemize}
    \item A systematic comparison of pooling strategies for Arabic text classification, demonstrating that simple mean pooling outperforms complex learned aggregation methods on limited training data.
    \item Observations about the dataset: human-written texts average 632 words versus 303 for machine-generated
    \item A training recipe with layer-wise learning rate decay and multi-sample dropout regularization
\end{itemize}

Our final system scores 0.75 F1 on the shared task test set.

\footnote{Code: \url{https://github.com/KickItLikeShika/abjadgeneval}}

\section{Background}

\subsection{Task Setup}
The AbjadEval task frames Arabic AI-generated text detection as a binary classification problem. Given an input text $x$, the system must predict a label $y \in \{\text{human}, \text{machine}\}$ indicating whether the text was written by a human or generated by an AI system.

\subsection{Dataset}
The competition dataset consists of 5,298 Arabic text samples with a balanced class distribution (50\% human, 50\% machine-generated). Table~\ref{tab:dataset} summarizes the dataset statistics.

\begin{table}[h]
\centering
\begin{tabular}{lcc}
\toprule
\textbf{Statistic} & \textbf{Human} & \textbf{Machine} \\
Samples & 2,649 & 2,649 \\
Avg. Words & 632.0 & 303.0 \\
Avg. Characters & 3,806.7 & 1,865.5 \\
Max Words & 3,068 & 1,969 \\
\bottomrule
\end{tabular}
\caption{Dataset statistics by class. Human-written texts are significantly longer than machine-generated texts.}
\label{tab:dataset}
\end{table}

A notable characteristic of the dataset is the substantial length difference between classes: human-written texts average 632 words compared to only 303 words for machine-generated texts. This suggests that text length could be a discriminative feature, though our model learns to capture more nuanced patterns.

\subsection{Related Work}
Previous approaches to AI-generated text detection have employed statistical methods \cite{gehrmann2019gltr}, fine-tuned language models, and watermarking techniques \cite{kirchenbauer2023watermark}. For Arabic specifically, transformer-based models like AraBERT \cite{arabert} and CAMeL-BERT \cite{camelbert} have shown strong performance on various NLP tasks. Recent work on multilingual text embeddings, particularly the E5 family \cite{e5}, has demonstrated excellent cross-lingual transfer capabilities.

More recent work has shifted toward fine-tuning language models directly for detection. The intuition is simple: if a model like BERT \cite{transformers} can learn what "natural" text looks like during pre-training, it should also be able to learn what generated text looks like with supervised fine-tuning. DetectGPT \cite{detectgpt} took a different approach, using perturbation-based methods that don't require any training data at all—though these zero-shot methods typically lag behind supervised ones when labeled data is available.

\section{System Overview}
\subsection{Model Architecture}
Our system is built on the multilingual E5-large encoder \cite{e5}, which consists of 24 transformer layers with a hidden size of 1,024. We add a classification head on top of the pooled representations.

The architecture follows a standard encoder-classifier setup. Input text is first tokenized and passed through the E5-large encoder, which produces a contextualized representation for each token. These token-level representations are then aggregated into a single fixed-size vector using a pooling operation. Finally, this pooled vector passes through a classification head that outputs probabilities for each class (human or machine).

\subsection{Pooling Strategies}
We experimented with several pooling strategies before settling on mean pooling:
\textbf{Mean Pooling:} The simplest approach, where we average the hidden states across all non-padded tokens. Each token contributes equally to the final representation.

\textbf{Weighted Layer Pooling:} Instead of using only the final transformer layer, this method learns to combine outputs from multiple layers. The intuition is that different layers capture different types of information: lower layers tend to encode surface-level features while higher layers capture more semantic content. We assign a learnable weight to each layer, and take a weighted average, with weights normalized using softmax.

\textbf{Multi-Head Attention Pooling:} Rather than treating all tokens equally, this approach learns which tokens to focus on. We use 8 learnable query vectors, each attending to the token sequence independently. The resulting 8 context vectors are concatenated and projected back to the hidden dimension. 

\textbf{Gated Fusion:} When combining multiple pooling methods, we use learned sigmoid gates to control how much each pooling output contributes. The gates are computed from the concatenation of all pooling outputs, allowing the model to dynamically weight different representations based on the input.

\subsection{Classification Head}
The pooled representation passes through a feed-forward layer with layer normalization, GELU activation, and dropout before the final classifier.
We also use multi-sample dropout \cite{multisampledropout}: during training, we apply 5 different dropout masks (with rates 0.1, 0.15, 0.2, 0.25, and 0.3) and average the resulting logits. This acts like a small ensemble within a single forward pass, improving regularization without additional inference cost.

\subsection{Loss Function}
We use Focal Loss \cite{focalloss} instead of standard cross-entropy. Focal loss down-weights easy examples and focuses training on harder cases by scaling the loss based on prediction confidence.

\section{Experimental Setup}
\subsection{Data Split}
We trained on the full competition training set containing 5,298 samples. Due to the blind test evaluation setup.

\subsection{Hyperparameters}
Table~\ref{tab:hyperparams} details our training configuration.

\begin{table}[h]
\centering
\begin{tabular}{ll}
\toprule
\textbf{Parameter} & \textbf{Value} \\
\midrule
Model & multilingual-e5-large \\
Max Sequence Length & 512 tokens \\
Batch Size & 16 \\
Gradient Accumulation & 4 steps \\
Effective Batch Size & 64 \\
Learning Rate & $2 \times 10^{-5}$ \\
Weight Decay & 0.01 \\
LLRD Decay Factor & 0.95 \\
Epochs & 2 \\
Warmup Ratio & 10\% \\
Scheduler & Cosine with warmup \\
\bottomrule
\end{tabular}
\caption{Training hyperparameters.}
\label{tab:hyperparams}
\end{table}

\subsection{Layer-wise Learning Rate Decay}
We apply layer-wise learning rate decay (LLRD) to prevent catastrophic forgetting of pretrained knowledge. Lower transformer layers receive smaller learning rates.

\subsection{Other Implementation Details}
We utilized Dynamic Padding to ensure that sequences are padded to max length within each batch for efficiency.

\section{Results}
Our system achieved an \textbf{F1 score of 0.75} on the official test set using mean pooling with the E5-large encoder.

\subsection{Pooling Strategy Results}
Table~\ref{tab:ablation} presents our comparison of pooling strategies during development.

\begin{table}[h]
\centering
\begin{tabular}{lc}
\toprule
\textbf{Pooling Strategy} & \textbf{Test F1} \\
\midrule
Mean Pooling & \textbf{0.75} \\
Weighted Layer Pooling + Attention  \\ + Gated Fusion & 0.70 \\
Weighted Layer Pooling + Attention & 0.71 \\
\bottomrule
\end{tabular}
\caption{Pooling strategy comparison on development set (2 samples). All methods achieved perfect dev scores, but mean pooling performed best on test.}
\label{tab:ablation}
\end{table}

Mean pooling demonstrated superior generalization on the held-out test set. Complex pooling strategies with more learnable parameters showed signs of overfitting.

\subsection{Analysis: Why Mean Pooling Works Best}
We hypothesize that simple mean pooling outperformed complex aggregation strategies for several reasons:

\textbf{1. Limited Training Data:} With only 5,298 training samples, sophisticated pooling mechanisms like weighted layer pooling (which learns to weigh 20+ layer weights) and multi-head attention pooling (with learned query vectors and projection matrices) introduce many additional parameters that require substantial data to train effectively.

\textbf{2. Pretrained Model Quality:} The E5-large model already produces high-quality token representations. Mean pooling preserves these representations without introducing additional learned transformations that may degrade under limited supervision.

\textbf{3. Regularization:} Mean pooling acts as implicit regularization by not adding learnable parameters to the pooling stage. The classification signal must flow through the fixed aggregation, preventing the model from overfitting through complex pooling patterns.

\textbf{4. Distributional Robustness:} Mean pooling treats all tokens equally, which may be beneficial when the discriminative features are distributed throughout the text rather than concentrated in specific positions.

\subsection{Error Analysis}
Analysis of the dataset reveals that human-written texts are approximately twice as long as machine-generated texts (632 vs. 303 words on average). This length disparity could serve as a discriminative feature, but also poses challenges:

\begin{itemize}
    \item \textbf{Truncation Effects:} With a maximum sequence length of 512 tokens, longer human texts are truncated, potentially losing discriminative information.
    \item \textbf{Length Bias:} The model may partially rely on length as a proxy feature, which could reduce robustness to length-controlled adversarial examples.
\end{itemize}

\begin{figure}[h]
    \centering
    \includegraphics[width=\linewidth]{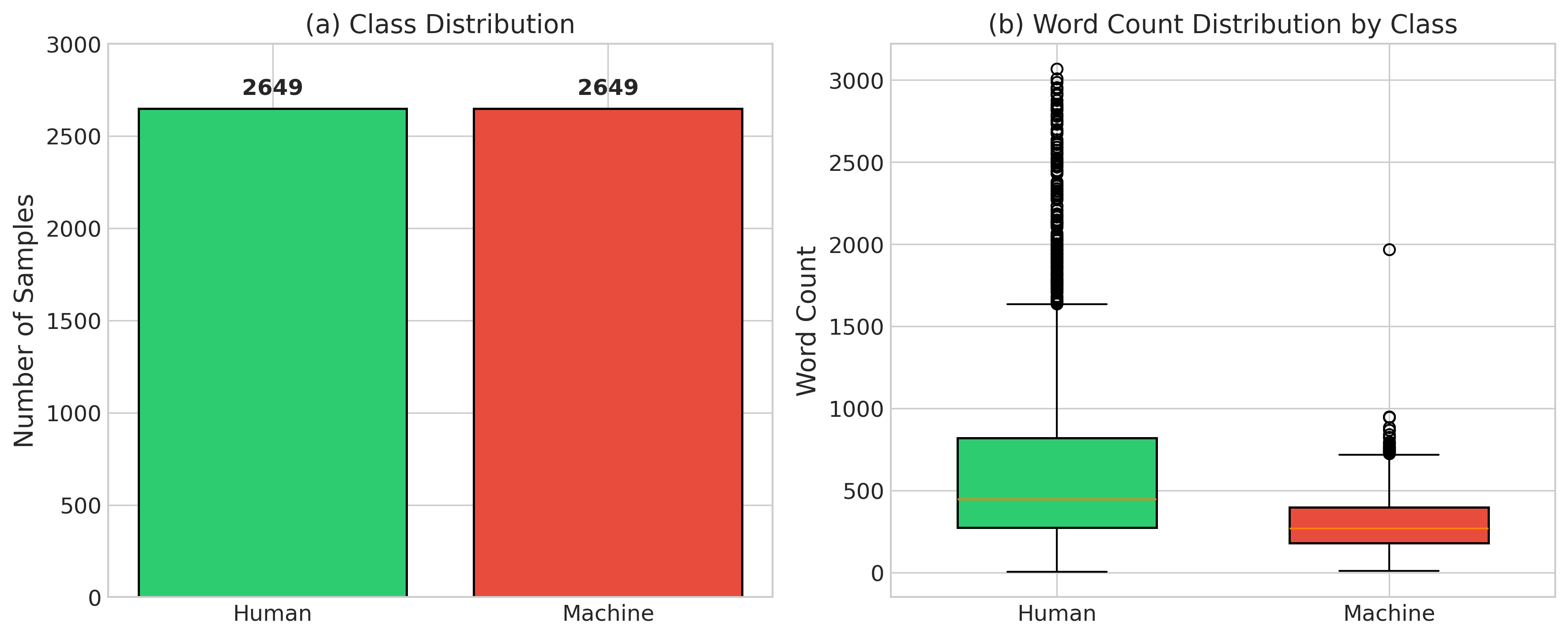}
    \caption{Dataset analysis: (a) Balanced class distribution with 2,649 samples per class. (b) Word count distribution showing human texts are significantly longer.}
    \label{fig:analysis}
\end{figure}

Figure~\ref{fig:analysis} shows the word count distribution by class, illustrating the clear separation between human and machine-generated texts.

\section{Conclusion}
We presented the our system for the AbjadEval Arabic AI-generated text shared-task, achieving an F1 score of 0.75 using the multilingual E5-large encoder with mean pooling. Our key finding is that simple mean pooling outperforms sophisticated aggregation strategies like weighted layer pooling and multi-head attention pooling when training data is limited.

\textbf{Limitations:} Our system was trained only on the provided competition data without external datasets.

\textbf{Future Work:} Investigating: (1) adding more data for training, (2) longer context windows to capture full document content, (3) ensemble methods combining multiple pooling strategies, and (4) the relationship between training data size and optimal pooling complexity.

\bibliography{custom}
\end{document}